\documentclass[conference]{IEEEtran}
\IEEEoverridecommandlockouts
\usepackage{color}
\usepackage{cite}
\usepackage{amsmath,amssymb,amsfonts}
\usepackage{algorithmic}
\usepackage{graphicx}
\usepackage{textcomp}
\usepackage{xcolor}
\usepackage{lipsum}
\usepackage{booktabs}
\usepackage{multirow}
\usepackage{algorithmic}
\usepackage{algorithm}
\usepackage[table]{xcolor}
\usepackage{pgfmath}
\usepackage{mathtools}
\usepackage{booktabs}
\usepackage{array}
\usepackage{hyperref}
\usepackage{etoolbox}
\makeatletter
\AfterEndEnvironment{algorithm}{\let\@algcomment\relax}
\AtEndEnvironment{algorithm}{\kern2pt\hrule\relax\vskip3pt\@algcomment}
\let\@algcomment\relax
\newcommand\algcomment[1]{\def\@algcomment{\footnotesize#1}}
\renewcommand\fs@ruled{\def\@fs@cfont{\bfseries}\let\@fs@capt\floatc@ruled
  \def\@fs@pre{\hrule height.8pt depth0pt \kern2pt}%
  \def\@fs@post{}%
  \def\@fs@mid{\kern2pt\hrule\kern2pt}%
  \let\@fs@iftopcapt\iftrue}
\makeatother

\usepackage[table]{xcolor} 
\definecolor {bestcolor}{HTML}{35b779} 
\definecolor {secondcolor}{HTML}{6dcd59} 
\definecolor {thirdcolor}{HTML}{b4de2c} 
\newcommand{\best}[1]{\cellcolor{bestcolor}{#1}}
\newcommand{\second}[1]{\cellcolor{secondcolor}{#1}}
\newcommand{\third}[1]{\cellcolor{thirdcolor}{#1}}
\usepackage{listings}
\usepackage[dvipsnames]{xcolor}

\def\BibTeX{{\rm B\kern-.05em{\sc i\kern-.025em b}\kern-.08em
    T\kern-.1667em\lower.7ex\hbox{E}\kern-.125emX}}
\begin{document}

\title{PointRFT: Explicit Reinforcement Fine-tuning for Point Cloud Few-shot Learning}

\author{Yankai Wang, Yiding Sun, Qirui Wang, Pengbo Li, Chaoyi Lu and Dongxu Zhang 
\thanks{Yankai Wang, Yiding Sun, Qirui Wang, Chaoyi Lu and Dongxu Zhang are with the School of Software Engineering, Xi'an Jiaotong University, Xi'an 710048, China; (e-mail: zhangdongxu@stu.xjtu.edu.cn) \textit{(Corresponding author: Dongxu Zhang).}}
\thanks{Yankai Wang and Pengbo Li are also with the International School, Beijing University of Posts and Telecommunications, Beijing 100876, China.}
}

\maketitle

\begin{abstract}
Understanding spatial dynamics and semantics in point cloud is fundamental for comprehensive 3D comprehension. While reinforcement learning algorithms such as Group Relative Policy Optimization (GRPO) have recently achieved remarkable breakthroughs in large language models by incentivizing reasoning capabilities through strategic reward design, their potential remains largely unexplored in the 3D perception domain. This naturally raises a pivotal question: Can RL-based methods effectively empower 3D point cloud fine-tuning? In this paper, we propose PointRFT, the first reinforcement fine-tuning paradigm tailored specifically for point cloud representation learning. We select three prevalent 3D foundation models and devise specialized accuracy reward and dispersion reward functions to stabilize training and mitigate distribution shifts. Through comprehensive few-shot classification experiments comparing distinct training paradigms, we demonstrate that PointRFT consistently outperforms vanilla supervised fine-tuning (SFT) across diverse benchmarks. Furthermore, when organically integrated into a hybrid Pretraining-SFT-RFT paradigm, the representational capacity of point cloud foundation models is substantially unleashed, achieving state-of-the-art performance particularly under data-scarce scenarios.
\end{abstract}


\section{Introduction}
\label{sec:intro}
Grasping spatial dynamics and semantics within point clouds serves as the cornerstone for holistic 3D understanding. This necessitates continuous development by the community of pretrained foundation models with enhanced generalization and representation capabilities. These exquisite models prove instrumental for a spectrum of downstream tasks, encompassing reasoning~\cite{zhang2025ascot,zhang2026chain,zhang2026notallqueries,wang2026personalq}, object classification~\cite{chen2025uni,prcv,sun6064487curve3d}, semantic segmentation~\cite{SUN2026112800,lan2026reco,lan2026performance}, action recognition~\cite{che2025lemon}, autonomous driving~\cite{zhang2026igasa,du2023degradationtgrs,10298249tnnls,ZHANG2026133318nc}, and embodied robotics~\cite{sun2026align}. 

Conventional point cloud perception methods typically depend on generative or contrastive pretrained models for feature extraction, succeeded by task-specific heads and supervised fine-tuning (SFT). We denote this pipeline as the `Pretraining-SFT' paradigm (Pre-S). Although demonstrating effectiveness on certain benchmarks~\cite{zhang2026pointcot,gong2025med}, these paradigms exhibit constrained cross-task and cross-domain generalization and frequently necessitate dedicated models per dataset or task, thereby impeding scalability and flexibility. Recently, drawing inspiration from Deepseek-R1~\cite{guo2025deepseek}, numerous studies prove that reinforcement learning (RL) coupled with Group Relative Policy Optimization (GRPO) enhances multimodal large language model (MLLM) performance~\cite{liu2025visual,yu2025perception,you2025seg}. These developments highlight the notable efficacy and expansive potential of RL-based fine-tuning (Pre-R) across diverse visual tasks. Consequently, we naturally pose the following question:

\textbf{Can RL-based methods effectively empower 3D point cloud fine-tuning?}

\begin{figure}[t]
\centering
\includegraphics[width=1.0\columnwidth]{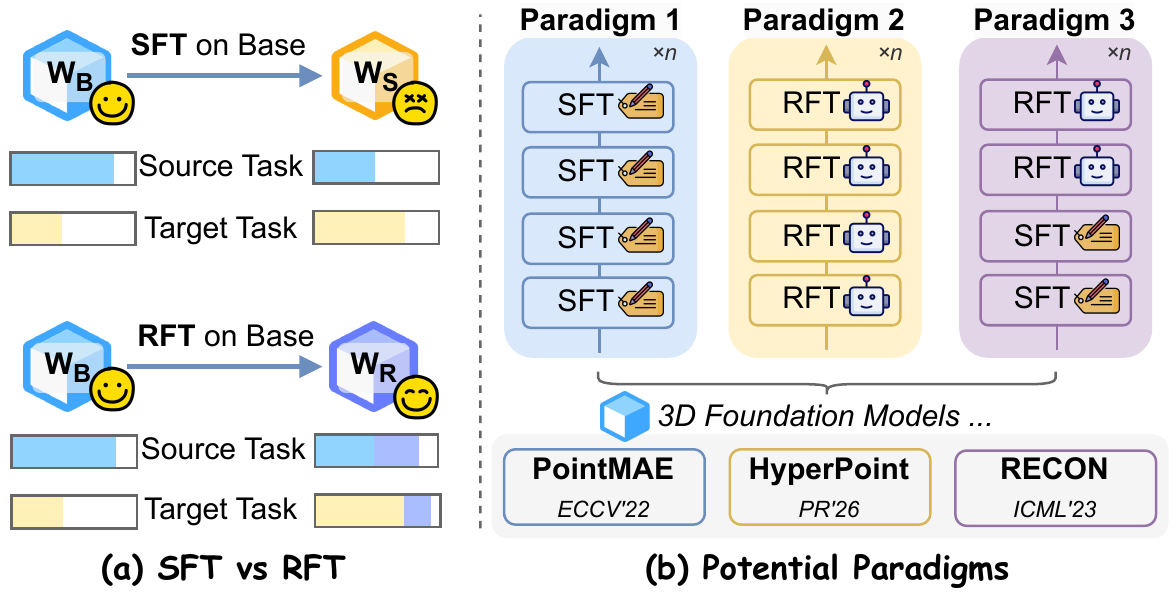}
\caption{\textbf{Schematic illustration of our PointRFT.} (a) SFT frequently provokes catastrophic forgetting, particularly when knowledge is transferred across domains. Reinforcement fine-tuning mitigates this decay and equips the backbone with broader generalization. (b) Grounded in this behavior, we introduce three reusable paradigms: Pre-S, Pre-R, and Pre-S-R, and benchmark them on multiple few-shot classification tasks using three prevalent 3D foundation models.}
\label{fig1}
\end{figure}

Regrettably, to our best knowledge, the community still lacks algorithmic breakthroughs in this area. This gap stems from differing feedback mechanisms inherent to RL-based algorithms in Large Language Models (LLM) versus point cloud foundation model. In contrast to language, point clouds are grounded in physical reality and harbor definitive objective ground truth, \textit{e.g.} labels, points, or bounding boxes. Such characteristic conflicts with the implicit exploratory nature of GRPO algorithms. More crucially, as depicted in Fig.~\ref{fig1}(a), SFT-based methods prove more readily applicable than RL-based methods, albeit suffering from overfitting on limited point cloud datasets under rigid supervision. Indeed, existing RL-based methods largely confine themselves to a narrow visual understanding scope, typically reasoning tasks like image QA~\cite{huang2025vision}, captioning~\cite{li2025videochat}, and action localization~\cite{wu2025tempr1}.

To address these limitations, we introduce PointRFT, a novel framework that substitutes vanilla SFT with RL-based methods in 3D perception. We replace GRPO's typically implicit and subjective reward function with an explicit, quantifiable metric. Furthermore, we leverage accuracy reward and dispersion reward to mitigate distribution shift during fine-tuning. Following~\cite{xu2025grpo}, we evaluate three representative point cloud foundation models across two experimental settings, encompassing the three paradigms illustrated in Fig.~\ref{fig1}(b). Our  results reveal that SFT suffers from dataset overfitting and catastrophic forgetting under few-shot scenarios. Meanwhile, experimental evidence demonstrates that PointRFT not only alleviates these issues but also effectively instills the fundamentals of point cloud classification in models. Upon shifting to the `Pretraining-SFT-RFT' (Pre-S-R) paradigm, we fully unleash the potential of foundation models and attain state-of-the-art performance in the base-to-new scenario. 

Our contributions can be summarized as follows:

\begin{itemize}
    \item We introduce PointRFT for the first time, a simple yet effective RL-based paradigm enabling object-level few-shot classification in point cloud perception.
    \item We design specialized explicit reward functions for point cloud perception tasks that promote knowledge acquisition rather than SFT-style memorization.
    \item Through comprehensive experiments across two settings, we systematically compare the positive impacts of RFT, SFT, and hybrid paradigms on point cloud perception, paving the way for further community research.
\end{itemize}

\section{Related Work}
\subsection{Pre-S Paradigm in 3D Perception}
Within the domain of point cloud representation learning, unsupervised pretraining strategies are principally divided into contrastive and generative paradigms. Contrastive methods~\cite{ijcv,guo2026momentummemoryknowledgedistillation,GUO2025109039} strive to maximize agreement among diverse augmentations of identical point clouds while concurrently minimizing similarity across disparate samples. PointContrast~\cite{PointContrast2020} marks a seminal milestone, subsequently advanced by MaskPoint~\cite{liu2022masked} with a binary classification task that discriminates masked object points from noise. ReCon~\cite{qi2023contrast} integrates contrastive objectives within generative pipelines to alleviate overfitting in original architecture. HyperPoint~\cite{SUN2026112800} successfully extends this pipeline into hyperbolic space, further strengthening model representation capabilities through explicit constraints. 

On the other hand, generative methods~\cite{che2025stitch} aim to reconstruct point clouds from latent representations by encoding inputs into a feature space and subsequently decoding them. Building upon MAE~\cite{he2022masked}, Point-MAE~\cite{pang2022masked} streamlines training and furnishes a mature solution for the research community. Thereafter, multimodal integration alongside recent architectures from other domains continues to enrich representations. Joint-MAE~\cite{guo2023joint} captures 2D-3D interactions, while PointGPT~\cite{chen2024pointgpt} adapts the GPT framework to point clouds. PointDif~\cite{zheng2024point} and PointDico~\cite{li2025pointdico} employ diffusion models as an alternative to MAE for pretraining. 

With powerful foundation models in place, incremental training via SFT becomes straightforward. However, despite high annotation costs, SFT is still considered to merely mimic predefined answers rather than discover what works best. We shift the training paradigm from data scaling in SFT to the strategic design of reward functions tailored to specific 3D vision tasks.

\subsection{RFT Paradigm in MLLM Perception}
Building upon the proven effectiveness of reasoning in LLM, a rising trend of works seeks to transfer this capability to MLLMs, thereby facilitating reasoning across diverse visual tasks. Vision-R1~\cite{huang2025vision} addresses intricate image reasoning within visual question answering. Perception-R1~\cite{yu2025perception} subsequently broadens this paradigm to encompass image object detection, thereby uncovering the potential of RL for perception-centric tasks. Seg-R1~\cite{you2025seg} proposes a decoupled reasoning-segmentation framework leveraging GRPO-based RL to produce explicit chain-of-thought reasoning alongside positional prompts for image segmentation. VideoChat-R1~\cite{li2025videochat} implements reinforcement fine-tuning across three spatio-temporal tasks to improve perception and reasoning in video understanding. GRPO-RM~\cite{xu2025grpo} validates the efficacy of integrating GRPO into DINO~\cite{oquabdinov2} backbones, yielding substantial enhancements in both visual and semantic logic understanding. 

Although these methods have attained remarkable progress in reasoning, they exclusively support either image or video reasoning. In this work, we take a pioneering step toward applying RL to point cloud perception, aiming to bridge this gap and advance spatial understanding.

\begin{figure*}[t]
\centering
\includegraphics[width=1.0\linewidth]{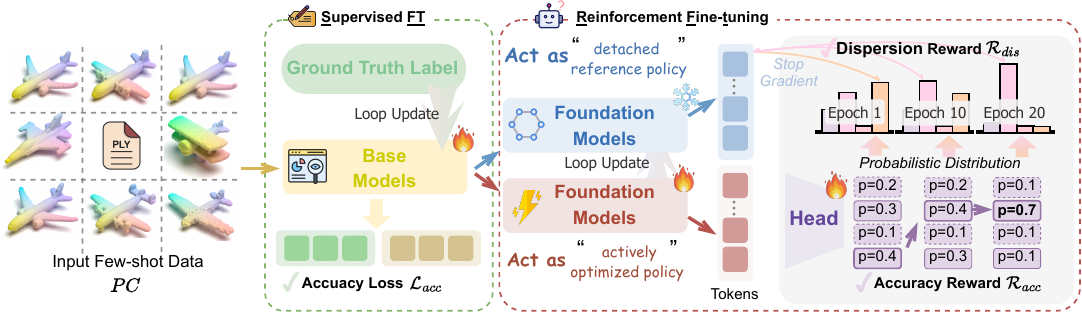}
\caption{\textbf{Framework of PointRFT.} Following the vanilla point cloud fine-tuning pipeline, the input point cloud is fed into a pre-trained {\color[RGB]{205,201,66} base model} and fine-tuned via Eq.~\ref{eq1}. For RL fine-tuning, we treat the base model before the update as the detached reference policy and the {\color[RGB]{160,78,58}updated base model} as the actively optimized policy to stabilize training. Building on this, we propose PointRFT, which trains the foundation model to maximize reward via Alg.~\ref{alg:code}. At each epoch, the parameters of the {\color[RGB]{67,114,176}old model} are updated and incorporated into the loss without gradient backflow. Since RFT and SFT share the same labeled data and inputs, the two paradigms do not conflict. In other words, we can maximize the downstream performance of the base model by first applying SFT and then RFT.}
\label{fig2}
\end{figure*}

\section{Method}
In this section, we draw inspiration from GRPO to formulate a RFT paradigm tailored for point cloud perception. To curb possible distribution shift during training, we devise customized accuracy and dispersion rewards. Subsequently, we delineate three compatible training schemes that synergistically leverage SFT and RFT to enhance few-shot learning capabilities. The overall framework is illustrated in Fig.~\ref{fig2}.
\subsection{Preliminaries}
Following Deepseek-R1~\cite{guo2025deepseek}, we adopt GRPO as the RL algorithm for optimization. Unlike SFT, which optimizes models through token-level losses, RL-based methods like GRPO employ policy gradients computed from reward loss for optimization. Such formulation encourages reasoning through exploration of a substantially larger solution space. Without loss of generality, let $Q$ denote the question set, $o$ be the label for a question $q$, and $\pi_{\theta_{\text {ref}}}$ represent the frozen reference model. SFT algorithms aim to optimize model $\pi_\theta$ via the following objective:
\begin{equation}
\begin{split}
\mathcal{L}_{\mathrm{}}(\theta) &= -\mathbb{E}_{q \sim Q, o \sim \pi_{\theta_{\text{ref}}}(\cdot \mid q)}\left[\log \pi_\theta(o \mid q)\right] \\
&= -\mathbb{E}_{q \sim Q, o \sim \pi_{\theta_{\text{ref}}}(\cdot \mid q)}\left[\sum_{t=1}^{|o|} \log \pi_\theta\left(o_t \mid q, o_{<t}\right)\right].
\end{split}
\label{eq1}
\end{equation}

\begin{algorithm}[t]
\caption{Pseudocode of PointRFT in a PyTorch-like style.}
\label{alg:code}
\definecolor{codeblue}{RGB}{0,102,204}
\newcommand{\comcol}{\color{codeblue}}
\lstset{
  backgroundcolor=\color{white},
  basicstyle=\fontsize{8pt}{8pt}\ttfamily\selectfont,
  columns=fullflexible,
  breaklines=true,
  captionpos=b,
  commentstyle=\fontsize{8pt}{8pt}\color{codeblue},
  keywordstyle=\fontsize{8pt}{8pt},
}
\begin{lstlisting}[language=python,mathescape=true]
# x: input point cloud
# y: corresponding labels
# m: batch
# e: training epoch
# h: projection head
# a, b: hyperparameter

$\pi_{\theta_{old}}$ = $\pi_{\theta_{base}}$ + h # build the initial model
for epoch in range(e): 
    $\pi_{\theta_{old}}\leftarrow\pi_\theta$ # copy the old model
    for m in loader: # load a minibatch 
        probs = $\pi_{\theta_{old}}$(x) # forward propagation
        r_acc = accuracy_reward(y_true, num_classes) # accuracy reward
        r_dis = -probs # dispersion reward
        rewards = a * r_acc + b * r_dis

        advantages = (rewards - mean(rewards))/std(rewards) 

        ratio = probs / old_probs # policy gradient
        clipped_ratio = clip(ratio, 1-$\varepsilon$, 1+$\varepsilon$) # distribution maintaining
        loss = -min(ratio * advantages, clipped_ratio * advantages).mean()

        loss.backward()
        optimizer.step() 
\end{lstlisting}
\end{algorithm}

Conversely, let $\pi_{\theta_{\text{old}}}$ denote the policy model and $\left\{o_1, o_2,\cdots, o_G\right\}$ be a group of responses from $\pi_{\theta_{\text {old}}}$ for question $q$. Given the absence of a reference model in representation learning, we omit the Kullback-Leibler (KL) divergence regularizer~\cite{xu2025grpo}. GRPO algorithms aim to optimize model $\pi_\theta$ through the following objective:
\begin{equation}
\begin{aligned}
\mathcal{J}_{\mathrm{}}(\theta)
& =\mathbb{E}_{q \sim Q,\left\{o_i\right\}_{i=1}^G \sim \pi_{\theta_{\text {old}}}} \left[\frac{1}{G} \sum_{i=1}^G \min \left(\frac{\pi_\theta\left(o_i \mid q\right)}{\pi_{\theta_{\text{old}}}\left(o_i \mid q\right)} A_i,\right.\right. \\
&\left.\left. \operatorname{clip}\left(\frac{\pi_\theta\left(o_i \mid q\right)}{\pi_{\theta_{\text {old}}}\left(o_i \mid q\right)}, 1-\epsilon, 1+\epsilon\right) A_i\right)\ \right],
\end{aligned}
\end{equation}
where $\varepsilon$ denotes the clipping hyperparameter. Let $\left\{r_1, r_2, \ldots, r_G\right\}$ be the group reward, the advantage $A_i$ is obtained through intra-group standardization:
\begin{equation}
A_i=\frac{r_i-\operatorname{mean}\left(\left\{r_1, r_2, \ldots, r_G\right\}\right)}{\operatorname{std}\left(\left\{r_1, r_2, \ldots, r_G\right\}\right)}.
\end{equation}

\begin{table*}[t]
\centering
\caption{Few-shot learning result on the ScanObjectNN Dataset. We conduct this analysis across three distinct splits for both 5-way 1-shot and 5-way 5-shot scenarios. The performance metrics are reported as accuracy percentages, accompanied by their respective standard deviations. The darker the color, the higher the result.}
\resizebox{\linewidth}{!}{
\begin{tabular}{lccccccc}
\toprule
\multirow{3}{*}{\textbf{Method}} & \multirow{3}{*}{\textbf{Paradigm}}\qquad\qquad\qquad & \multicolumn{6}{c}{\textbf{ScanObjectNN}}                                                    \\ \cline{3-8}
                                 &                                    & \multicolumn{3}{c}{\textbf{5-way 1-shot}}              & \multicolumn{3}{c}{\textbf{5-way 5-shot}}              \\ \cline{3-8}
                                 &                                    & Split 1          & Split 2          & Split 3          & Split 1          & Split 2          & Split 3          \\ \hline
\multirow{5}{*}{Point-MAE~\cite{pang2022masked}\qquad}
& Pre-S (10\%)\qquad\qquad\qquad
& 27.85 $\pm$ 0.62\qquad & 29.91 $\pm$ 0.68\qquad & 33.62 $\pm$ 0.87\qquad&63.77 $\pm$ 0.83&65.42 $\pm$ 1.37&67.08 $\pm$ 0.45 \\
& Pre-S\qquad\qquad\qquad
& 41.08 $\pm$ 1.93\qquad & 45.56 $\pm$ 1.04\qquad & 49.33 $\pm$ 1.12\qquad
&70.19 $\pm$ 1.94&72.64 $\pm$ 1.05&74.55 $\pm$ 1.26 \\
& Pre-R (10\%)\qquad\qquad\qquad
& \third{28.03 $\pm$ 0.91}\qquad & \third{29.76 $\pm$ 0.38}\qquad & \third{36.54 $\pm$ 0.87}\qquad&\best{67.08 $\pm$ 0.45}&\best{67.33 $\pm$ 1.72}&\best{68.91 $\pm$ 0.91}\\
& Pre-R\qquad\qquad\qquad
& \second{42.00 $\pm$ 1.59}\qquad & \second{47.22 $\pm$ 0.70}\qquad & \third{50.68 $\pm$ 1.44}\qquad
& \third{72.64 $\pm$ 1.05}\qquad & \best{76.25 $\pm$ 0.67}\qquad & \third{77.48 $\pm$ 1.59}\qquad \\
& Pre-S-R\qquad\qquad\qquad
& \third{48.95 $\pm$ 0.83}\qquad & \second{56.11 $\pm$ 0.66}\qquad & \best{61.50 $\pm$ 1.52}\qquad
& \third{74.13 $\pm$ 0.92}\qquad & \second{76.82 $\pm$ 1.21}\qquad & \second{78.50 $\pm$ 0.39}\qquad \\ \midrule
\multirow{5}{*}{HyperPoint~\cite{SUN2026112800}}
& Pre-S (10\%)\qquad\qquad\qquad
& 30.68 $\pm$ 0.44\qquad & 31.27 $\pm$ 0.39\qquad & 38.84 $\pm$ 1.61\qquad&63.27 $\pm$ 1.88&65.96 $\pm$ 0.74&67.15 $\pm$ 1.03 \\
& Pre-S\qquad\qquad\qquad
& 34.78 $\pm$ 0.99\qquad & 37.22 $\pm$ 0.70\qquad & 49.20 $\pm$ 0.84\qquad
&68.04 $\pm$ 0.56\qquad & 70.73 $\pm$ 0.82\qquad & 74.09 $\pm$ 1.47\qquad \\
& Pre-R (10\%)\qquad\qquad\qquad
& \second{29.33 $\pm$ 0.79}\qquad & \second{37.67 $\pm$ 0.41}\qquad & \best{40.09 $\pm$ 0.95}\qquad&\third{60.45 $\pm$ 1.27}&\second{65.83 $\pm$ 0.62}&\second{66.19 $\pm$ 1.96}\\
& Pre-R\qquad\qquad\qquad
& \third{38.14 $\pm$ 1.58}\qquad & \third{46.89 $\pm$ 0.72}\qquad & \second{51.74 $\pm$ 1.37}\qquad
& \second{73.50 $\pm$ 0.35}\qquad & \third{74.77 $\pm$ 1.58}\qquad & \third{77.08 $\pm$ 1.04}\qquad \\
& Pre-S-R\qquad\qquad\qquad
& \second{49.20 $\pm$ 0.84}\qquad & \third{47.46 $\pm$ 1.63}\qquad & \third{56.81 $\pm$ 1.27}\qquad
& \second{74.31 $\pm$ 1.83}\qquad & \third{74.92 $\pm$ 0.91}\qquad & \third{77.24 $\pm$ 0.73}\qquad \\ \midrule
\multirow{5}{*}{ReCon~\cite{qi2023contrast}}
& Pre-S (10\%)\qquad\qquad\qquad
& 31.74 $\pm$ 0.37\qquad & 35.09 $\pm$ 0.74\qquad & 40.14 $\pm$ 1.31\qquad&64.66 $\pm$ 0.49&66.11 $\pm$ 1.12&65.27 $\pm$ 0.87 \\
& Pre-S\qquad\qquad\qquad
& 40.47 $\pm$ 0.73\qquad & 47.85 $\pm$ 0.62\qquad & 51.19 $\pm$ 0.45\qquad
& 68.00 $\pm$ 1.45\qquad & 69.54 $\pm$ 0.56\qquad & 72.38 $\pm$ 1.71\qquad \\
& Pre-R (10\%)\qquad\qquad\qquad
& \best{32.05 $\pm$ 0.73}\qquad & \best{39.44 $\pm$ 1.96}\qquad & \second{38.92 $\pm$ 0.35}\qquad&\second{63.95 $\pm$ 0.94}&\third{65.19 $\pm$ 1.33}&6\third{5.62 $\pm$ 0.68}\\
& Pre-R\qquad\qquad\qquad
& \best{43.67 $\pm$ 0.54}\qquad & \best{50.11 $\pm$ 0.62}\qquad & \best{57.29 $\pm$ 1.83}\qquad
& \best{75.78 $\pm$ 0.91}\qquad & \second{75.52 $\pm$ 1.04}\qquad & \best{79.09 $\pm$ 1.92}\qquad \\
& Pre-S-R\qquad\qquad\qquad
& \best{56.90 $\pm$ 1.12}\qquad & \best{58.76 $\pm$ 0.87}\qquad & \best{61.54 $\pm$ 1.45}\qquad
& \best{77.34 $\pm$ 1.25}\qquad & \best{76.88 $\pm$ 0.67}\qquad & \best{79.23 $\pm$ 0.49}\qquad \\ \bottomrule
\end{tabular}}
\label{tab1}
\end{table*}

\begin{table}[t]
\caption{Few-shot learning result on the ModelNet40 Dataset. We conduct this analysis for both 10-way 10-shot and 10-way 20-shot scenarios. We conduct further experiments that use 10\% of the training time for comparison.}
\centering
\resizebox{\linewidth}{!}{
\begin{tabular}{lccc}
\toprule
\multirow{2}{*}{\textbf{Method}}&\multirow{2}{*}{\textbf{Paradigm}}\quad&\multicolumn{2}{c}{\textbf{ModelNet40}}\\\cline{3-4}
&&\textbf{10-way 10-shot}&\textbf{10-way 20-shot}\\\midrule
\multirow{5}{*}{Point-MAE~\cite{pang2022masked}}
&Pre-S (10\%)\quad&76.6 $\pm$ 4.3&79.9 $\pm$ 2.1\\
&Pre-S\quad&92.6 $\pm$ 4.1&95.0 $\pm$ 3.0\\
&Pre-R (10\%)\quad&\best{83.6 $\pm$ 5.3}&\best{84.9 $\pm$ 2.1}\\
&Pre-R\quad&\third{92.3 $\pm$ 4.5}&\third{95.0 $\pm$ 3.0}\\
&Pre-S-R\quad&\second{94.2 $\pm$ 3.3}&\third{95.6 $\pm$ 2.8}\\\midrule
\multirow{5}{*}{HyperPoint~\cite{SUN2026112800}}
&Pre-S (10\%)\quad&76.8 $\pm$ 2.3&76.7 $\pm$ 2.2\\
&Pre-S\quad&93.3 $\pm$ 4.0&95.6 $\pm$ 2.8\\
&Pre-R (10\%)\quad&\second{77.0 $\pm$ 1.8}&\second{78.3 $\pm$ 1.3}\\
&Pre-R\quad&\second{92.6 $\pm$ 3.7}&\second{95.5 $\pm$ 3.0}\\
&Pre-S-R\quad&\third{93.3 $\pm$ 3.9}&\second{95.7 $\pm$ 3.0}\\\midrule
\multirow{5}{*}{ReCon~\cite{qi2023contrast}}
&Pre-S (10\%)\quad&77.0 $\pm$ 1.9&78.7 $\pm$ 1.2\\
&Pre-S\quad&94.4 $\pm$ 1.9&94.9 $\pm$ 0.9\\
&Pre-R (10\%)\quad&\third{76.8 $\pm$ 1.8}&\second{78.3 $\pm$ 1.4}\\
&Pre-R\quad&\best{94.1 $\pm$ 4.4}&\best{95.9 $\pm$ 2.2}\\
&Pre-S-R\quad&\best{95.2 $\pm$ 0.8}&\best{96.5 $\pm$ 0.6}\\\bottomrule
\end{tabular}}
\label{tab2}
\end{table}

\begin{table}[t]
\caption{Few-shot learning result on the ShapeNetCore Dataset. We conduct this analysis for both 10-way 1-shot and 10-way 5-shot scenarios. We conduct further experiments that use 10\% of the training time for comparison.}
\centering
\resizebox{\linewidth}{!}{
\begin{tabular}{lccc}
\toprule
\multirow{2}{*}{\textbf{Method}}&\multirow{2}{*}{\textbf{Paradigm}}\quad&\multicolumn{2}{c}{\textbf{ShapeNetCore}}\\\cline{3-4}
&&\textbf{10-way 1-shot}&\textbf{10-way 5-shot}\\\midrule
\multirow{5}{*}{Point-MAE~\cite{pang2022masked}}
&Pre-S (10\%)\quad&56.08 $\pm$ 0.85&68.27 $\pm$ 1.36\\
&Pre-S\quad&59.83 $\pm$ 0.54&72.45 $\pm$ 1.78\\
&Pre-R (10\%)\quad&\best{53.91 $\pm$ 0.42}&\second{66.18 $\pm$ 1.09}\\
&Pre-R\quad&\third{66.70 $\pm$ 0.87}&\best{71.52 $\pm$ 1.63}\\
&Pre-S-R\quad&\third{70.34 $\pm$ 0.95}&\third{73.09 $\pm$ 1.21}\\\midrule
\multirow{5}{*}{HyperPoint~\cite{SUN2026112800}}
&Pre-S (10\%)\quad&54.00 $\pm$ 0.33&63.88 $\pm$ 1.57\\
&Pre-S\quad&60.25 $\pm$ 0.69&69.46 $\pm$ 1.92\\
&Pre-R (10\%)\quad&\second{52.14 $\pm$ 0.76}&\third{65.37 $\pm$ 1.44}\\
&Pre-R\quad&\second{67.80 $\pm$ 0.58}&\second{71.23 $\pm$ 1.11}\\
&Pre-S-R\quad&\second{73.55 $\pm$ 0.99}&\second{74.72 $\pm$ 1.68}\\\midrule
\multirow{5}{*}{ReCon~\cite{qi2023contrast}}
&Pre-S (10\%)\quad&52.47 $\pm$ 1.15&68.93 $\pm$ 0.72\\
&Pre-S\quad&57.66 $\pm$ 1.88&73.25 $\pm$ 0.45\\
&Pre-R (10\%)\quad&\third{51.18 $\pm$ 1.63}&\best{70.54 $\pm$ 0.91}\\
&Pre-R\quad&\best{68.79 $\pm$ 1.37}&\third{71.02 $\pm$ 0.58}\\
&Pre-S-R\quad&\best{75.41 $\pm$ 1.74}&\best{76.33 $\pm$ 0.89}\\\bottomrule
\end{tabular}}
\label{tab3}
\end{table}

By eliminating the critic model and estimating relative advantages, GRPO attains markedly higher computational efficiency~\cite{guo2025deepseek}.

\subsection{Reward Design}
Once the advantages computation is established, we tailor the reward signal to suit point cloud representation learning. Consistent with~\cite{xu2025grpo}, we posit that similar instances form compact clusters in the embedding space. Nevertheless, given the discriminative nature of classification, the overall embedding distribution should remain sufficiently dispersion. These two objectives should be dynamically reweighted according to dataset statistics and label cardinality. Consequently, PointRFT’s reward is decomposed into an accuracy term and a dispersion term.

\textbf{Accuracy Reward.} Mirroring the original GRPO pipeline, we encourage the model to emit correct predictions. Yet point clouds only supply data–label pairs and lack the chain-of-thought annotations available in (M)LLMs, the indispensable group of responses $\left\{o_1, o_2, \cdots, o_G\right\}$ is absent. We therefore seek a surrogate that still yields rich output variation. The class-probability vector produced by the softmax layer serves this purpose. Unlike auto-regressive language models, we directly draw samples from this distribution. Concretely, for a $c$-class dataset $\left\{C_1, C_2, \cdots, C_c\right\}$, an input point cloud $x_k$ of true class $C_k$ gives rise to a probability mass that we treat as a pseudo-response ensemble. Under this scheme, the accuracy reward for any draw can be compactly expressed as:

\begin{equation}
\left\{\begin{array}{rl}
r_{\mathrm{acc}_i}=0 & i \neq k, \\
r_{\mathrm{acc}_i}=c & i=k .
\end{array}\right.
\end{equation}

Owing to the denser reward landscape of RL fine-tuning compared with SFT, we clamp the per-sample accuracy reward $r_{\operatorname{acc}_k}$ to a constant $c$ while enforcing $\mathbb{E} r_{\operatorname{acc}_i}=1$ to preserve training stability.

\textbf{Dispersion Reward.} HyperPoint~\cite{SUN2026112800} highlights that representation quality hinges on both alignment and dispersion. We further argue that point cloud perception presupposes two conditions: intra-class cohesion and global dispersion. However, with cross-entropy alone, vanilla SFT secures only weak alignment and leaves dispersion unattended. In chasing maximal accuracy, the model collapses features onto their centroids, inflating inter-class similarity and squandering embedding capacity. This geometric shrinkage is masked when data abound, yet emerges as a dominant error source under few-shot or out-of-domain shifts. To counteract this, the dispersion reward penalizes incorrect predictions:

\begin{equation}
r_{\text {dis}_i}=-p_i, \quad j \in\{1,2, \cdots, c\}.
\end{equation}

Intuitively, removing the dispersion reward collapses PointRFT back into vanilla SFT. In other words, dispersion is therefore the continuous, differentiable bridge that turns discrete labels into a learnable reward landscape. Once both components are established, the final reward is computed via summation, that is,
\begin{equation}
r_i=a*r_{\mathrm{acc}_i}+b*r_{\mathrm{dis}_i},
\label{er6}
\end{equation}
where $a$ and $b$ are hyperparameters that regulate training stability across datasets with varying label counts.

\subsection{Training Paradigms}
Fig.~\ref{fig2} illustrates that SFT and RFT are intrinsically complementary. Exhaustive comparisons already span linguistics, 2D vision and MLLM perception, yet the 3D domain remains largely uncharted. This omission stems from an implicit, LLM-centric bias, leaving a conspicuous gap in how these paradigms reshape visual representations. We outline the meta-objective $\mathcal{F}(\theta;\Phi)$ and three candidate training paradigms for future inquiry:

\begin{equation}
\mathcal{F}(\theta ; \Phi) \triangleq \begin{cases}\mathcal{L}_{\mathrm{SFT}}(\theta), & \Phi=\mathrm{SFT} \\ \mathcal{J}_{\mathrm{RFT}}(\theta), & \Phi=\mathrm{RFT} \\ \mathcal{H}(\theta), & \Phi=\mathrm{Hybrid} \end{cases}
\end{equation}
where $\Phi \in\{\mathrm{SFT}, \mathrm{RL}, \mathrm{Hybrid}\}$ switches the training regime. $\mathcal{L}_{\mathrm{SFT}}(\theta)$ and $\mathcal{J}_{\mathrm{RFT}}(\theta)$ denote the supervised and reinforcement losses respectively. The hybrid target $\mathcal{H}(\theta)$ is the bi-level optimization:
\begin{equation}
\mathcal{H}(\theta) \triangleq \mathcal{J}_{\mathrm{GRPO}}(\theta) \quad \text{\textit{s.t.}} \quad \theta \in \arg \min _{\theta^{\prime}} \mathcal{L}_{\mathrm{SFT}}\left(\theta^{\prime}\right).
\end{equation}

\begin{table}[t]
\centering
\caption{\textbf{Comparison in terms of accuracy with different dispersion reward function and hyperparameter settings.} ``set to 0 " means only an accuracy reward is applied. 10-way 10-shot, 10-way 1-shot, and 5-way 1-shot are tested respectively.}
\resizebox{\linewidth}{!}{
\begin{tabular}{lccc}
\toprule
\textbf{Dataset}\qquad             & \textbf{Epsilon}\qquad          & \textbf{Dispersion Reward}                    & \textbf{Accuracy} \\ \midrule
\multirow{3}{*}{ModelNet40~\cite{wu20153d}}\qquad & \multirow{3}{*}{$\varepsilon=0.2$}\qquad   & set to 0                               & 83.4 $\pm$ 4.9   \\
                             & & a=1,b=1                                & 89.2 $\pm$ 3.3   \\
                             &   & Eq.~\ref{er6}   & 92.6 $\pm$ 3.7   \\ \midrule
\multirow{3}{*}{ShapeNetCore~\cite{chang2015shapenet}}\qquad  & \multirow{3}{*}{$\varepsilon=0.3$}\qquad  & set to 0                              & 56.12 $\pm$ 1.93    \\
                            &  & a=1,b=1                                & 60.66 $\pm$ 1.89    \\
                             &  & Eq.~\ref{er6}   & 65.42 $\pm$ 1.15    \\ \midrule
\multirow{3}{*}{ScanObjectNN~\cite{uy2019revisiting}}\qquad  & \multirow{3}{*}{$\varepsilon=0.5$}\qquad & set to 0                             & 33.22 $\pm$ 1.34    \\
                             &   & a=1,b=1                               & 35.91 $\pm$ 0.73    \\
                            &  & Eq.~\ref{er6}   & 37.04 $\pm$ 1.58    \\ \bottomrule
\end{tabular}}
\label{ab}
\end{table}

\section{Experiments}
\subsection{Experiment Settings}
We evaluate PointRFT on three baselines: Point-MAE~\cite{pang2022masked}, HyperPoint~\cite{SUN2026112800}, and ReCon~\cite{qi2023contrast}. Each baseline was pretrained on ShapeNet~\cite{chang2015shapenet} for 300 epochs. Unless otherwise specified, all baselines adhere to the original configurations detailed in their respective papers. For consistency, we fine-tune these models using a supervised protocol, aligning network settings with those of RFT. For PointRFT, the hyperparameters are set as $\varepsilon=0.2, a=1$, and $b=2$. Across all datasets, the Adam optimizer is utilized to train the models. Each epoch consists of 400 meta-training and 600 validation episodes, with samples randomly drawn from the training set. Each meta-training episode is configured as an $N$-way, $M$-shot setup, where $N$ classes are selected, each with $M$ examples. After meta-training, we conduct final evaluations using 700 meta-testing episodes, randomly sampled from the test set, maintaining the $N$-way, $M$-shot configuration. The final performance is assessed by computing the mean classification accuracy and standard deviation across these meta-testing episodes. All experiments are conducted on NVIDIA A100-SXM4-80GB GPUs.

\begin{figure}[t]
\centering
\includegraphics[width=1.0\columnwidth]{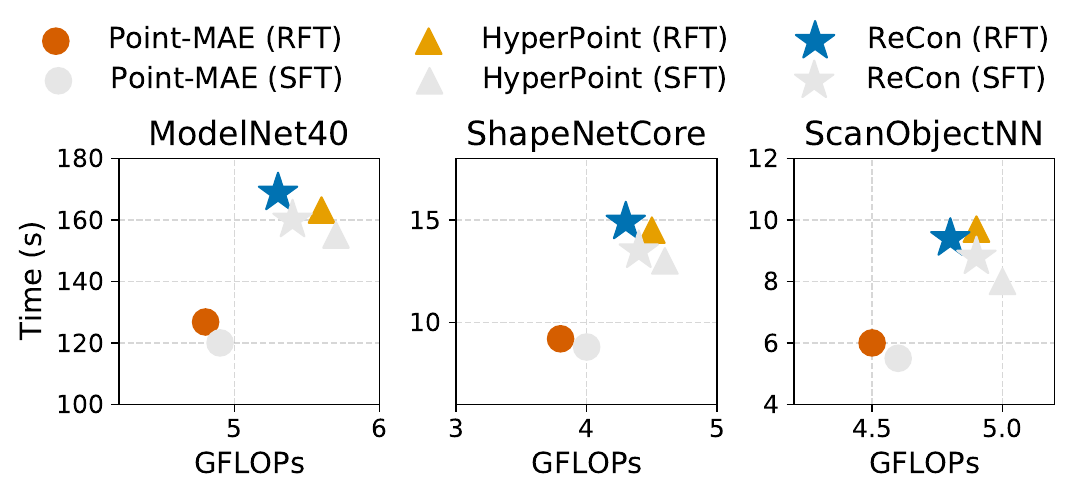}
\caption{\textbf{Comparison of computational costs in different datasets.} We report the GLOPs and training time (s) for each epoch. 10-way 10-shot, 10-way 1-shot and 5-way 1-shot are selected respectively.}
\label{fig3}
\end{figure}

\subsection{Classification: ModelNet40}
\textbf{Dataset.} ModelNet40~\cite{wu20153d} serves as a standard benchmark for few-shot 3D object recognition, assembling 40 CAD-based categories. Its episodic splits require a classifier to label query shapes from only a handful of support examples per class.

\textbf{Comparison Results.} Empirical validation spans three backbones using ModelNet40 10-way, 10-shot and 10-way, 20-shot scenarios. Results are consolidated in Table~\ref{tab2}. PointRFT marginally underperforms vanilla SFT, indicating that ample support examples erase the advantage of reinforcement fine-tuning. Meanwhile, multimodal pre-training~\cite{SUN2026112800,qi2023contrast} consistently boosts accuracy over its unimodal counterpart~\cite{pang2022masked}, an edge that widens once the few-shot support set becomes sufficiently large.

\subsection{Classification: ShapeNetCore}
\textbf{Dataset.} ShapeNetCore~\cite{chang2015shapenet} extends the evaluation to 70 CAD classes, offering broader semantic coverage. The benchmark gauges whether models can reliably categorize sparse point clouds when each category is represented by a small, fixed support set.

\textbf{Comparison Results.} Table~\ref{tab3} benchmarks RFT against competing paradigms on ShapeNetCore under 10-way-1-shot and 10-way-5-shot scenarios, probing behavior across increasing supervision scarcity. PointRFT surpasses all counterparts in every regime, evidencing its suitability for 3D few-shot tasks. Equipped with the proposed Pre-S-R enhancement, the model unlocks additional capacity, delivering gains of +10.51\%, +13.30\% and +17.75\% over the vanilla SFT baselines.

\subsection{Classification: ScanObjectNN}
\textbf{Dataset.} ScanObjectNN~\cite{uy2019revisiting} shifts the assessment from pristine CAD data to real-world scans. Acquired via depth sensors, it injects nuisances such as background clutter, partial occlusion, and unknown scale, thereby testing few-shot generalization under the distributional shifts common in practice. The collection provides 15 annotated object categories.

\textbf{Comparison Results.} We contrast RFT with SFT on ScanObjectNN under 5-way 1-shot and 5-way 5-shot scenarios across three disjoint splits~\cite{ye2023closer}. Each split guarantees that test point clouds are semantically isolated from training data, thereby measuring strict generalization within the prescribed budget. In Table~\ref{tab1}, PointRFT consistently ranks first in every split, confirming its robustness to real-world nuisances. Pre-S-R further amplifies this edge under 5-way-5-shot, echoing the trend previously observed on Table~\ref{tab3}.

\subsection{Ablation Studies}
\textbf{Hyperparameters.} We probe hyperparameter sensitivity in two ablations. Using HyperPoint~\cite{SUN2026112800} as the backbone, Table~\ref{ab} sweeps the weights $a$ and $b$. Disabling dispersion reward ($b$ = 0) hurts accuracy, corroborating its role as a soft regularizer. On the other hand, large $\varepsilon$ destabilizes optimization, a symptom of sparse RL updates that favors over-fitting in low-data regimes, a pathology also observed in SFT. A moderate clip is therefore mandatory to restrain the effective parameter drift.

\textbf{Computational Burden.} Fig.~\ref{fig3} quantifies the computational footprint of PointRFT using two metrics: per-epoch training time and GFLOPs. Although PointRFT demands slightly more resources than SFT, this overhead is offset by faster convergence and superior accuracy, rendering the trade-off acceptable.

\section{Conclusion}
In this paper, we propose PointRFT, a pioneering reinforcement fine-tuning paradigm that extends the success of GRPO-driven optimization from language modeling to the realm of 3D point cloud perception. By substituting the implicit rewards with explicit, task-oriented accuracy and dispersion signals, we mitigate the distribution shift and over-fitting endemic to scarce-data supervised fine-tuning. Our three-pronged evaluation framework, encompassing Pretraining-SFT, Pretraining-RFT, and the hybrid Pretraining-SFT-RFT paradigms, provides the first systematic investigation into how reinforcement learning reshapes geometric representation learning. Despite rigorous variable control and ablations, limited compute and time have prevented us from extending PointRFT to additional 3D perception tasks such as semantic segmentation or action recognition, leaving its broader efficacy yet to be confirmed.  In the future, we will further disentangle the precise conditions under which RFT departs from SFT when adapting point cloud backbones.




\bibliographystyle{IEEEbib}
\bibliography{icme2026_template_anonymized}

@article{chen2024pointgpt,
  title={Pointgpt: Auto-regressively generative pre-training from point clouds},
  author={Chen, Guangyan and Wang, Meiling and Yang, Yi and Yu, Kai and Yuan, Li and Yue, Yufeng},
  journal={Proc. Adv. Neural Inf. Process. Syst.},
  volume={36},
  year={2024}
}

@inproceedings{PointContrast2020,
    author = {Saining Xie and Jiatao Gu and Demi Guo and Charles R. Qi and Leonidas Guibas Or Litany},
    title = {PointContrast: Unsupervised Pre-training for 3D Point Cloud Understanding},
    booktitle = {Eur. Conf. Comput. Vis.},
    year = {2020},
}

@inproceedings{he2022masked,
  title={Masked autoencoders are scalable vision learners},
  author={He, Kaiming and Chen, Xinlei and Xie, Saining and Li, Yanghao and Doll{\'a}r, Piotr and Girshick, Ross},
  booktitle={Proc. IEEE Conf. Comput. Vis. Pattern Recognit.},
  pages={16000--16009},
  year={2022}
}

@inproceedings{pang2022masked,
  title={Masked autoencoders for point cloud self-supervised learning},
  author={Pang, Yatian and Wang, Wenxiao and Tay, Francis EH and Liu, Wei and Tian, Yonghong and Yuan, Li},
  booktitle={Eur. Conf. Comput. Vis.},
  pages={604--621},
  year={2022},
}

@inproceedings{liu2022masked,
  title={Masked discrimination for self-supervised learning on point clouds},
  author={Liu, Haotian and Cai, Mu and Lee, Yong Jae},
  booktitle={Eur. Conf. Comput. Vis.},
  pages={657--675},
  year={2022},
  organization={Springer}
}

@inproceedings{qi2023contrast,
  title={Contrast with Reconstruct: Contrastive 3D Representation Learning Guided by Generative Pretraining},
  author={Qi, Zekun and Dong, Runpei and Fan, Guofan and Ge, Zheng and Zhang, Xiangyu and Ma, Kaisheng and Yi, Li},
  booktitle={Int. Conf. Mach. Learn.},
  year={2023}
}

@article{li2025pointdico,
  title={PointDico: Contrastive 3D Representation Learning Guided by Diffusion Models},
  author={Li, Pengbo and Sun, Yiding and Cheng, Haozhe},
  journal={arXiv preprint arXiv:2512.08330},
  year={2025}
}

@inproceedings{zheng2024point,
  title={Point Cloud Pre-training with Diffusion Models},
  author={Zheng, Xiao and Huang, Xiaoshui and Mei, Guofeng and Hou, Yuenan and Lyu, Zhaoyang and Dai, Bo and Ouyang, Wanli and Gong, Yongshun},
  booktitle={Proc. IEEE/CVF Conf. Comput. Vis. Pattern Recognit.},
  pages={22935--22945},
  year={2024}
}

@inproceedings{guo2023joint,
  title={Joint-MAE: 2D-3D Joint Masked Autoencoders for 3D Point Cloud Pre-training},
  author={Guo, Ziyu and Zhang, Renrui and Qiu, Longtian and Li, Xianzhi and Heng, Pheng-Ann},
  booktitle={Proc. Int. Joint Conf. Artif. Intell.},
  pages={791--799},
  year={2023}
}

@article{chang2015shapenet,
  title={Shapenet: An information-rich 3d model repository},
  author={Chang, Angel X and Funkhouser, Thomas and Guibas, Leonidas and Hanrahan, Pat and Huang, Qixing and Li, Zimo and Savarese, Silvio and Savva, Manolis and Song, Shuran and Su, Hao and others},
  journal={arXiv preprint arXiv:1512.03012},
  year={2015}
}

@article{ye2023closer,
  title={A Closer Look at Few-Shot 3D Point Cloud Classification},
  author={Ye, Chuangguan and Zhu, Hongyuan and Zhang, Bo and Chen, Tao},
  journal={Int. J. Comput. Vis.},
  volume={131},
  number={3},
  pages={772--795},
  year={2023}
}

@article{lan2026reco,
  title={ReCo-KD: Region-and Context-Aware Knowledge Distillation for Efficient 3D Medical Image Segmentation},
  author={Lan, Qizhen and Hsu, Yu-Chun and Khan, Nida Saddaf and Jiang, Xiaoqian},
  journal={arXiv preprint arXiv:2601.08301},
  year={2026}
}

@article{chen2025uni,
  title={Uni-NTFM: A Unified Foundation Model for EEG Signal Representation Learning},
  author={Chen, Zhisheng and Zhang, Yingwei and Lan, Qizhen and Liu, Tianyu and Wang, Huacan and Ding, Yi and Jia, Ziyu and Chen, Ronghao and Wang, Kun and Zhou, Xinliang},
  journal={arXiv preprint arXiv:2509.24222},
  year={2025}
}

@article{lan2026performance,
  title={From Performance to Practice: Knowledge-Distilled Segmentator for On-Premises Clinical Workflows},
  author={Lan, Qizhen and Choi, Aaron and Ma, Jun and Wang, Bo and Zhao, Zhaogming and Jiang, Xiaoqian and Hsu, Yu-Chun},
  journal={arXiv preprint arXiv:2601.09191},
  year={2026}
}

@article{zhang2025ascot,
  title={Ascot: An adaptive self-correction chain-of-thought method for late-stage fragility in llms},
  author={Zhang, Dongxu and Yang, Ning and Sun, Yiding and Zhu, Jihua and Yang, Jinnan and Xin, Miao and Tian, Baoliang},
  journal={arXiv preprint arXiv:2508.05282},
  year={2025}
}

@article{zhang2026chain,
  title={Chain-of-Thought Compression Should Not Be Blind: V-Skip for Efficient Multimodal Reasoning via Dual-Path Anchoring},
  author={Zhang, Dongxu and Sun, Yiding and Tan, Cheng and Yan, Wenbiao and Yang, Ning and Zhu, Jihua and Zhang, Haijun},
  journal={arXiv preprint arXiv:2601.13879},
  year={2026}
}

@article{zhang2026pointcot,
  title={PointCoT: A Multi-modal Benchmark for Explicit 3D Geometric Reasoning},
  author={Zhang, Dongxu and Sun, Yiding and Li, Pengcheng and Liu, Yumou and Lin, Hongqiang and Xu, Haoran and Mu, Xiaoxuan and Lin, Liang and Yan, Wenbiao and Yang, Ning and others},
  journal={arXiv preprint arXiv:2602.23945},
  year={2026}
}

@article{sun2026align,
  title={Align then Adapt: Rethinking Parameter-Efficient Transfer Learning in 4D Perception},
  author={Sun, Yiding and Zhu, Jihua and Cheng, Haozhe and Lu, Chaoyi and Yang, Zhichuan and Chen, Lin and Wang, Yaonan},
  journal={IEEE Trans. Multimedia},
  year={2026}
}

@InProceedings{prcv,
author="Han, Xingguang
and Sun, Yiding
and Lu, Chaoyi",
title="Rethinking Regressor in 3D Gaussian Pretraining",
booktitle="Pattern Recognit. Comput. Vis.",
year="2026",
pages="177--190",
}

@article{zhang2026notallqueries,
      title={Not All Queries Need Deep Thought: CoFiCot for Adaptive Coarse-to-fine Stateful Refinement}, 
      author={Zhang, Dongxu and Lin, Hongqiang and Sun, Yiding and Wang, Pengyu and Wang, Qirui and Yang, Ning and Zhu, Jihua},
      journal={arXiv preprint arXiv:2603.08251},
      year={2026}
}

@article{zhang2026igasa,
  title={IGASA: Integrated Geometry-Aware and Skip-Attention Modules for Enhanced Point Cloud Registration},
  author={Zhang, Dongxu and Zhu, Jihua and Li, Shiqi and Yan, Wenbiao and Xu, Haoran and Fan, Peilin and Lu, Huimin},
  journal={IEEE Trans. Circuits Syst. Video Technol.},
  year={2026},
  publisher={IEEE}
}

@article{che2025lemon,
  title={Lemon: A large endoscopic monocular dataset and foundation model for perception in surgical settings},
  author={Che, Chengan and Wang, Chao and Vercauteren, Tom and Tsoka, Sophia and Garcia-Peraza-Herrera, Luis C},
  journal={arXiv preprint arXiv:2503.19740},
  year={2025}
}

@article{che2025stitch,
  title={A Stitch in Time: Learning Procedural Workflow via Self-Supervised Plackett-Luce Ranking},
  author={Che, Chengan and Wang, Chao and Chen, Xinyue and Tsoka, Sophia and Garcia-Peraza-Herrera, Luis C},
  journal={arXiv preprint arXiv:2511.17805},
  year={2025}
}

@article{gong2025med,
  title={Med-CMR: A Fine-Grained Benchmark Integrating Visual Evidence and Clinical Logic for Medical Complex Multimodal Reasoning},
  author={Gong, Haozhen and Ji, Xiaozhong and Liu, Yuansen and Wu, Wenbin and Yan, Xiaoxiao and Liu, Jingjing and Wu, Kai and Pan, Jiazhen and Jian, Bailiang and Zhang, Jiangning and others},
  journal={arXiv preprint arXiv:2512.00818},
  year={2025}
}

@article{ijcv,
title = "Unsupervised Hyperspectral Image Super-Resolution via Self-Supervised Modality Decoupling",
author = "Songcheng Du and Yang Zou and Zixu Wang and Xingyuan Li and Ying Li and Changjing Shang and Qiang Shen",
year = "2026",
month = jan,
day = "9",
journal = "Int. J. Comput. Vis.",
issn = "0920-5691",
}

@article{du2023degradationtgrs,
  title={Degradation aware unfolding network for spectral super-resolution},
  author={Du, Songcheng and Leng, Yihong and Liang, Xinyi and Li, Jiaojiao and Liu, Wei and Du, Qian},
  journal={IEEE Geosci. Remote Sens. Lett.},
  volume={21},
  pages={1--5},
  year={2023},
  publisher={IEEE}
}

@ARTICLE{10298249tnnls,
  author={Li, Jiaojiao and Du, Songcheng and Song, Rui and Li, Yunsong and Du, Qian},
  journal={IEEE Trans. Neural Netw. Learn. Syst.}, 
  title={Progressive Spatial Information-Guided Deep Aggregation Convolutional Network for Hyperspectral Spectral Super-Resolution}, 
  year={2025},
  volume={36},
  number={1},
  pages={1677--1691},
  doi={10.1109/TNNLS.2023.3325682}
}

@article{GUO2025109039,
title = {BPMambaMIL: A bio-inspired prototype-guided multiple instance learning for oncotype DX risk assessment in histopathology},
journal = {Computer Methods and Programs in Biomedicine},
volume = {272},
pages = {109039},
year = {2025},
author = {Yongxin Guo and Ziyu Su and Onur C. Koyun and Hao Lu and Robert Wesolowski and Gary Tozbikian and M. Khalid Khan Niazi and Metin N. Gurcan},
}

@misc{guo2026momentummemoryknowledgedistillation,
      title={Momentum Memory for Knowledge Distillation in Computational Pathology}, 
      author={Yongxin Guo and Hao Lu and Onur C. Koyun and Zhengjie Zhu and Muhammet Fatih Demir and Metin Nafi Gurcan},
      year={2026},
      eprint={2602.21395},
      archivePrefix={arXiv},
      primaryClass={cs.CV},
      url={https://arxiv.org/abs/2602.21395}, 
}

@article{ZHANG2026133318nc,
title = {CMHANet: A cross-modal hybrid attention network for point cloud registration},
journal = {Neurocomputing},
volume = {680},
pages = {133318},
year = {2026},
issn = {0925-2312},
author = {Dongxu Zhang and Yingsen Wang and Yiding Sun and Haoran Xu and Peilin Fan and Jihua Zhu},
}

@article{sun6064487curve3d,
  title={Curve3D: Curvature-Aware Masked Autoencoder for Self-supervised Point Cloud Understanding},
  author={Sun, Yiding and Lu, Chaoyi and Cheng, Haozhe and Wang, Jun and Lu, Huimin and Chen, Lin and Zhu, Jihua},
  journal={Available at SSRN 6064487},
  year ={2026}
}

@article{wang2026personalq,
  title   = {PersonalQ: Select, Quantize, and Serve Personalized Diffusion Models for Efficient Inference},
  author  = {Qirui Wang and Qi Guo and Yiding Sun and Junkai Yang and Dongxu Zhang and Shanmin Pang and Qing Guo},
  journal = {arXiv preprint arXiv:2603.22943},
  year    = {2026},
  url     = {https://arxiv.org/abs/2603.22943}
}

@inproceedings{uy2019revisiting,
  title={Revisiting point cloud classification: A new benchmark dataset and classification model on real-world data},
  author={Uy, Mikaela Angelina and Pham, Quang-Hieu and Hua, Binh-Son and Nguyen, Thanh and Yeung, Sai-Kit},
  booktitle={Proc. IEEE/CVF Int. Conf. Comput. Vis.},
  pages={1588--1597},
  year={2019}
}

@inproceedings{wu20153d,
  title={3d shapenets: A deep representation for volumetric shapes},
  author={Wu, Zhirong and Song, Shuran and Khosla, Aditya and Yu, Fisher and Zhang, Linguang and Tang, Xiaoou and Xiao, Jianxiong},
  booktitle={Proc. IEEE Conf. Comput. Vis. Pattern Recognit.},
  pages={1912--1920},
  year={2015}
}

@article{SUN2026112800,
  title={HyperPoint: Multimodal 3D foundation model in hyperbolic space},
  author={Sun, Yiding and Cheng, Haozhe and Lu, Chaoyi and Li, Zhengqiao and Wu, Minghong and Lu, Huimin and Zhu, Jihua},
  journal={Pattern Recognit.},
  volume={173},
  pages={112800},
  year={2026}
}

@article{guo2025deepseek,
  title={Deepseek-r1: Incentivizing reasoning capability in llms via reinforcement learning},
  author={Guo, Daya and Yang, Dejian and Zhang, Haowei and Song, Junxiao and Zhang, Ruoyu and Xu, Runxin and Zhu, Qihao and Ma, Shirong and Wang, Peiyi and Bi, Xiao and others},
  journal={arXiv preprint arXiv:2501.12948},
  year={2025}
}

@article{liu2025visual,
  title={Visual-RFT: Visual Reinforcement Fine-Tuning},
  author={Liu, Ziyu and Sun, Zeyi and Zang, Yuhang and Dong, Xiaoyi and Cao, Yuhang and Duan, Haodong and Lin, Dahua and Wang, Jiaqi},
  journal={CoRR},
  year={2025}
}

@article{yu2025perception,
  title={Perception-r1: Pioneering perception policy with reinforcement learning},
  author={Yu, En and Lin, Kangheng and Zhao, Liang and Yin, Jisheng and Wei, Yana and Peng, Yuang and Wei, Haoran and Sun, Jianjian and Han, Chunrui and Ge, Zheng and others},
  journal={arXiv preprint arXiv:2504.07954},
  year={2025}
}

@article{you2025seg,
  title={Seg-R1: Segmentation Can Be Surprisingly Simple with Reinforcement Learning},
  author={You, Zuyao and Wu, Zuxuan},
  journal={arXiv preprint arXiv:2506.22624},
  year={2025}
}

@article{huang2025vision,
  title={Vision-r1: Incentivizing reasoning capability in multimodal large language models},
  author={Huang, Wenxuan and Jia, Bohan and Zhai, Zijie and Cao, Shaosheng and Ye, Zheyu and Zhao, Fei and Xu, Zhe and Hu, Yao and Lin, Shaohui},
  journal={arXiv preprint arXiv:2503.06749},
  year={2025}
}

@article{oquabdinov2,
  title={DINOv2: Learning Robust Visual Features without Supervision},
  author={Oquab, Maxime and Darcet, Timoth{\'e}e and Moutakanni, Th{\'e}o and Vo, Huy V and Szafraniec, Marc and Khalidov, Vasil and Fernandez, Pierre and Haziza, Daniel and Massa, Francisco and El-Nouby, Alaaeldin and others},
  journal={IEEE Trans. Mach. Learn. Res.},
  year={2023}
}

@article{li2025videochat,
  title={Videochat-r1: Enhancing spatio-temporal perception via reinforcement fine-tuning},
  author={Li, Xinhao and Yan, Ziang and Meng, Desen and Dong, Lu and Zeng, Xiangyu and He, Yinan and Wang, Yali and Qiao, Yu and Wang, Yi and Wang, Limin},
  journal={arXiv preprint arXiv:2504.06958},
  year={2025}
}

@article{wu2025tempr1,
  title={TempR1: Improving Temporal Understanding of MLLMs via Temporal-Aware Multi-Task Reinforcement Learning},
  author={Wu, Tao and Yang, Li and Zhan, Gen and Liao, Yiting and Li, Junlin and Fu, Deliang and Zhang, Li and Wang, Limin},
  journal={arXiv preprint arXiv:2512.03963},
  year={2025}
}

@article{xu2025grpo,
  title={GRPO-RM: Fine-Tuning Representation Models via GRPO-Driven Reinforcement Learning},
  author={Xu, Yanchen and Jiao, Ziheng and Zhang, Hongyuan and Li, Xuelong},
  journal={arXiv preprint arXiv:2511.15256},
  year={2025}
}

\end{document}